\documentclass{article}

\usepackage[preprint]{neurips_2026}
\workshoptitle{World Models in Physical AI}

\usepackage[utf8]{inputenc}
\usepackage[T1]{fontenc}
\usepackage{amsmath,amssymb,amsfonts}
\usepackage{booktabs}
\usepackage{graphicx}
\usepackage{microtype}
\usepackage{xcolor}
\usepackage{multirow}
\usepackage{tabularx}
\usepackage{float}
\usepackage{tikz}
\usetikzlibrary{arrows.meta,calc,positioning}
\usepackage{hyperref}
\hypersetup{
  hidelinks,
  pdftitle={In CEM, a World Model Is Also a Proposal Mechanism},
  pdfauthor={Oliver Obst and Frieder Stolzenburg}
}
\newcommand{\PTWalkerDInitial}{0.254}
\newcommand{\PTWalkerDFinal}{0.235}
\newcommand{\PTWalkerDContrastMinimum}{-0.035}
\newcommand{\PTWalkerDContrastMaximum}{-0.009}
\newcommand{\PTWalkerDMeanInitial}{0.200}
\newcommand{\PTWalkerDMeanFinal}{0.288}
\newcommand{\PTWalkerDLogWidthInitial}{0.293}
\newcommand{\PTWalkerDLogWidthFinal}{0.165}
\newcommand{\PTWalkerDLinearWidthInitial}{0.165}
\newcommand{\PTWalkerDLinearWidthFinal}{0.147}
\newcommand{\PTWalkerDWassersteinInitial}{0.260}
\newcommand{\PTWalkerDWassersteinFinal}{0.324}
\newcommand{\PTWalkerFirstActionDInitial}{0.349}
\newcommand{\PTWalkerFirstActionDFinal}{0.326}
\newcommand{\PTWalkerFirstActionGapRawInitial}{0.349}
\newcommand{\PTWalkerFirstActionGapRawFinal}{0.218}
\newcommand{\PTWalkerFirstActionGapNormalisedInitial}{0.998}
\newcommand{\PTWalkerFirstActionGapNormalisedFinal}{1.321}
\newcommand{\PTWalkerPairwiseAccuracyInitial}{0.503}
\newcommand{\PTWalkerPairwiseAccuracyFinal}{0.503}
\newcommand{\PTWalkerAdjustedElitePrecisionInitial}{0.000}
\newcommand{\PTWalkerAdjustedElitePrecisionFinal}{-0.001}
\newcommand{\PTWalkerDWassersteinPercentIncrease}{25}
\newcommand{\PTWalkerProposalWidthInitial}{0.350}
\newcommand{\PTWalkerProposalWidthFinal}{0.159}
\newcommand{\PTWalkerClippingDisplacementRawInitial}{0.116}
\newcommand{\PTWalkerClippingDisplacementRawFinal}{0.005}
\newcommand{\PTWalkerFloorOccupancyFinalPercent}{1.2}
\newcommand{\PTWalkerAttributionScorerPercent}{28.7}
\newcommand{\PTWalkerAttributionPoolPercent}{31.6}
\newcommand{\PTWalkerAttributionPairingPercent}{39.7}
\newcommand{\PTWalkerAttributionScorerExceedsPoolSeedCount}{2}
\newcommand{\PTWalkerAttributionSeedCount}{6}
\newcommand{\PTRandomNonlinearWalkerSeedZeroResidualFinal}{0.0023}
\newcommand{\PTRandomNonlinearWalkerSeedOneResidualFinal}{0.0021}
\newcommand{\PTRandomNonlinearWalkerSeedTwoResidualFinal}{0.0035}

\newcommand{\PTRandomNonlinearWalkerPositiveResidualCountFinal}{5}
\newcommand{\PTRandomNonlinearWalkerResidualSeedCount}{6}
\newcommand{\PTCheetahDInitial}{0.252}
\newcommand{\PTCheetahDFinal}{0.235}
\newcommand{\PTCheetahDContrastMinimum}{-0.023}
\newcommand{\PTCheetahDContrastMaximum}{-0.004}
\newcommand{\PTCheetahDMeanInitial}{0.184}
\newcommand{\PTCheetahDMeanFinal}{0.292}
\newcommand{\PTCheetahDLogWidthInitial}{0.303}
\newcommand{\PTCheetahDLogWidthFinal}{0.156}
\newcommand{\PTCheetahDLinearWidthInitial}{0.160}
\newcommand{\PTCheetahDLinearWidthFinal}{0.138}
\newcommand{\PTCheetahDWassersteinInitial}{0.244}
\newcommand{\PTCheetahDWassersteinFinal}{0.324}
\newcommand{\PTCheetahFirstActionDInitial}{0.261}
\newcommand{\PTCheetahFirstActionDFinal}{0.252}
\newcommand{\PTCheetahFirstActionGapRawInitial}{0.273}
\newcommand{\PTCheetahFirstActionGapRawFinal}{0.184}
\newcommand{\PTCheetahFirstActionGapNormalisedInitial}{0.779}
\newcommand{\PTCheetahFirstActionGapNormalisedFinal}{1.258}
\newcommand{\PTCheetahPairwiseAccuracyInitial}{0.535}
\newcommand{\PTCheetahPairwiseAccuracyFinal}{0.523}
\newcommand{\PTCheetahAdjustedElitePrecisionInitial}{0.071}
\newcommand{\PTCheetahAdjustedElitePrecisionFinal}{0.040}
\newcommand{\PTCheetahDWassersteinPercentIncrease}{33}
\newcommand{\PTCheetahProposalWidthInitial}{0.350}
\newcommand{\PTCheetahProposalWidthFinal}{0.131}
\newcommand{\PTCheetahClippingDisplacementRawInitial}{0.137}
\newcommand{\PTCheetahClippingDisplacementRawFinal}{0.006}
\newcommand{\PTCheetahFloorOccupancyFinalPercent}{1.4}
\newcommand{\PTCheetahAttributionScorerPercent}{66.2}
\newcommand{\PTCheetahAttributionPoolPercent}{10.0}
\newcommand{\PTCheetahAttributionPairingPercent}{23.8}
\newcommand{\PTCheetahAttributionScorerExceedsPoolSeedCount}{6}
\newcommand{\PTCheetahAttributionSeedCount}{6}
\newcommand{\PTRandomNonlinearCheetahSeedZeroResidualFinal}{0.0090}
\newcommand{\PTRandomNonlinearCheetahSeedOneResidualFinal}{0.0018}
\newcommand{\PTRandomNonlinearCheetahSeedTwoResidualFinal}{0.0050}

\newcommand{\PTRandomNonlinearCheetahPositiveResidualCountFinal}{5}
\newcommand{\PTRandomNonlinearCheetahResidualSeedCount}{6}

\newcommand{\PTSpliceSelectionWalkerFinalNormalisedMean}{-0.405}

\newcommand{\PTSpliceSelectionCheetahFinalNormalisedMean}{-0.912}

\newcommand{\PTSpliceSelectionFinalUnitCount}{6}

\newcommand{\PTSpliceHeldoutWalkerFinalNormalisedMean}{-0.253}

\newcommand{\PTSpliceHeldoutCheetahFinalNormalisedMean}{-0.620}

\newcommand{\PTSpliceHeldoutFinalUnitCount}{6}
\newcommand{\PTSpliceCombinedWalkerFinalNormalisedMean}{-0.329}
\newcommand{\PTSpliceCombinedWalkerFinalBestNormalisedMean}{-0.053}

\newcommand{\PTSpliceCombinedCheetahFinalNormalisedMean}{-0.766}
\newcommand{\PTSpliceCombinedCheetahFinalBestNormalisedMean}{-0.442}

\newcommand{\PTSpliceCombinedFinalUnitCount}{12}

\newcommand{\PTSpliceHeldoutStateOmissionCount}{4}

\newif\ifworkingdraft
\workingdraftfalse

\newif\ifarxivappendix
\arxivappendixtrue

\definecolor{draftblue}{RGB}{35,89,140}
\title{In CEM, a World Model Is Also a Proposal Mechanism}

\author{%
  Oliver Obst$^1$\thanks{\texttt{o.obst@unsw.edu.au}} \qquad\qquad
  Frieder Stolzenburg$^{2,1}$\thanks{\texttt{fstolzenburg@hs-harz.de}}\\[1em]
  $^1$UNSW Sydney \qquad $^2$Harz University of Applied Sciences
}

\begin{document}
\raggedbottom
\maketitle

\begin{abstract}
The cross-entropy method (CEM) uses world-model scores to select action
sequences and fit the distribution sampled in its next iteration. A scoring
error can therefore change both the present decision and the candidates
considered later. We evaluate these two roles separately. Four types of
predictive model generate CEM traces, and every model rescores every saved
candidate pool. Executing the same candidates in the environment provides a reference
elite set and proposal update.

Across twelve independently trained task-seed units on Walker and Cheetah,
the pre-specified proposal distance falls from the first to the final
CEM iteration in every unit.
Proposal widths contract and fitted means separate relative to the remaining
search width. Pairwise ranking agreement stays near chance on Walker and
declines on Cheetah; elite-set agreement does not improve. This comparison shows greater
variation between scorers than between pool sources on Cheetah; Walker has
variation in both and in their pairings. We use the original six units to
select Random nonlinear for a one-update intervention, without inspecting
intervention outcomes. Replacing its first model-ranked update with an
environment-ranked update lowers final realised selected-sequence cost in
those six units and in six further units held out from the selection.
\end{abstract}

\section{Introduction}

A world model used for planning determines which actions receive further
consideration. In model-predictive control (MPC), the planner predicts the
outcomes of candidate action sequences, selects a sequence, and executes its
first action before replanning. The cross-entropy method (CEM) performs this
search by repeatedly sampling candidates and fitting a new sampling
distribution to those with the lowest predicted
cost~\citep{bharadhwaj2020mpc}. We call that distribution the \emph{proposal}
and the retained candidates the \emph{elite set}.

As an illustrative example, consider two candidate pools for the same physical state and cost objective.
Pool A contains a sequence with realised cost 1, but the scorer selects one
with realised cost 5. This is a selection error: a lower-cost sequence was
available. Pool B contains only sequences with realised cost 8 or higher. Its scorer orders
every candidate correctly and selects the cost-8 sequence. Earlier CEM
updates could have left the planner searching this second region. Correct
ordering within that pool still produces a higher-cost choice than the
misranked choice from pool A.

A fixed-pool comparison with environment executions can measure ranking
errors, elite membership, proposal-refit disagreement, and selection regret
among the available candidates. In the example, it can expose the error in
pool A and confirm the ordering in pool B. It cannot show which lower-cost
sequences another proposal would have sampled. Correct within-pool ordering
does not ensure that the pool contains low-cost actions.

Evaluation also depends on where the candidates came from. Logged rollouts
use a data-collection policy; actor-centred ranking tests perturb a reference
action sequence. These tests can measure prediction or ordering on their
chosen action distributions. CEM changes its own distribution after each
elite decision. We call this \emph{planner-induced query shift}: model
scores change the optimiser's future queries. The physical system need not
change.

Rank agreement measured along a CEM trace already captures how ordering
changes during search~\citep{wang2026alignment}. Our comparison adds
pools produced by other models' update histories. Applying the same scorer
to those pools tests whether its disagreement depends on pool source;
applying different scorers to one pool compares their decisions on identical
actions. Realised candidate costs separately record which pools contain
lower-cost sequences.

Our comparison uses four predictors fitted over frozen Dreamer
contexts~\citep{hafner2020dreamer}. A context is the backbone's internal
representation of the observation history. Each predictor replaces the full
Dreamer dynamics during planning, so we call it a \emph{surrogate}. It
predicts future contexts under a candidate action sequence. A fixed readout
maps those contexts to the task variables used in the planning cost. The
four surrogates share data, transition form, cost, and CEM settings; their
feature maps differ.

Each surrogate first runs its own CEM search. We save the candidate pool at
every iteration, execute its candidates from the same restored simulator
state, and then score that pool with every surrogate. This separates two
roles: the \emph{generator} produced the pool through its earlier updates;
the \emph{scorer} ranks the saved pool now. We call this a \emph{crossed design}
because each of the four generators is paired with each of the four scorers,
giving all 16 combinations. A column follows one
scorer across pool sources. A row holds the pool fixed and compares scorers.
The matrix describes proposal disagreement associated with generators,
scorers, and their pairings. It does not assign causal effects to those
components.

\begin{figure}[t]
  \centering
  \resizebox{\textwidth}{!}{\begin{tikzpicture}[
  x=1cm,
  y=1cm,
  font=\large,
  separator/.style={draw=black!22, line width=0.45pt},
  flow/.style={draw=black!65, thick, -{Latex[length=2mm]}},
  model/.style={draw=orange!90!black, thick, -{Latex[length=2mm]}},
  env/.style={draw=green!45!black, thick, -{Latex[length=2mm]}},
  pool/.style={draw=blue!60, thick, fill=blue!8},
  dot/.style={circle, fill=blue!62, inner sep=1.25pt},
  smallnode/.style={draw=black!35, rounded corners=1pt, fill=white,
    align=center, inner sep=2.3pt, font=\normalsize},
]

\draw[separator] (2.68,-1.48) -- (2.68,1.58);
\draw[separator] (8.03,-1.48) -- (8.03,1.58);

\node[font=\bfseries] at (0,1.43) {One model-driven trace};
\node[font=\bfseries] at (5.35,1.43) {Crossed trace audit};
\node[font=\bfseries] at (10.70,1.43) {One-update elite splice};

\draw[pool] (-1.88,0.30) ellipse (0.63 and 0.43);
\foreach \p in {(-2.12,0.17),(-1.96,0.39),(-1.77,0.08),(-1.61,0.36)}
  \node[dot] at \p {};
\node[smallnode] (score) at (-0.55,0.30) {score\\$\widehat J_f$};
\node[smallnode] (elites) at (0.72,0.30) {retain\\$E^M$};
\draw[flow] (-1.23,0.30) -- (score.west);
\draw[model] (score.east) -- (elites.west);
\draw[model] (elites.east) -- (1.82,0.30);
\draw[pool] (1.86,0.30) ellipse (0.48 and 0.34);
\foreach \p in {(1.63,0.20),(1.78,0.41),(1.94,0.12),(2.08,0.36)}
  \node[dot] at \p {};
\node[font=\small, anchor=west] at (-2.32,-0.36) {pool from $q_i$};
\node[font=\small] at (1.86,0.82) {fit $q_{i+1}$};
\draw[model, rounded corners=3pt]
  (1.86,-0.05) -- (1.86,-0.75) -- (-2.78,-0.75) -- (-2.78,0.30)
  -- (-2.51,0.30);
\node[align=center, font=\normalsize] at (0,-1.22)
  {$q_{i+1}$ sets the centre and scale\\of the next query pool.};

\node[font=\normalsize, rotate=90] at (3.53,0.12) {pool source $g$};
\foreach \r in {0,1,2,3} {
  \node[font=\small] at (4.02,{0.78-0.48*\r}) {$g_{\the\numexpr\r+1\relax}$};
  \node[font=\small] at ({4.47+0.48*\r},1.12) {$f_{\the\numexpr\r+1\relax}$};
  \foreach \c in {0,1,2,3} {
    \pgfmathtruncatemacro{\diag}{\r==\c ? 1 : 0}
    \ifnum\diag=1
      \filldraw[draw=orange!85!black, fill=orange!22]
        ({4.25+0.48*\c},{0.56-0.48*\r}) rectangle ++(0.42,0.42);
    \else
      \filldraw[draw=blue!50, fill=blue!8]
        ({4.25+0.48*\c},{0.56-0.48*\r}) rectangle ++(0.42,0.42);
    \fi
  }
}
\node[font=\normalsize, text=orange!90!black, anchor=west] at (6.35,0.56)
  {self trace};
\node[font=\normalsize, text=blue!65!black, anchor=west] at (6.35,0.10)
  {cross score};
\node[align=center, font=\normalsize] at (5.35,-1.25)
  {Rows: pool source; columns: scorer.\\Environment costs are reused.};

\node[smallnode] (a0) at (8.84,0.35) {common\\pool $\mathcal A_0$};
\node[smallnode, fill=orange!8] (am) at (10.42,0.78) {$\mathcal A_1^M$};
\node[smallnode, fill=green!7] (ae) at (10.42,-0.08) {$\mathcal A_1^E$};
\draw[model] (a0.east) -- (am.west);
\draw[env] (a0.east) -- (ae.west);
\node[font=\small, text=orange!90!black] at (9.56,0.97)
  {$(\mu_1^M,\sigma_1^M)$};
\node[font=\small, text=green!45!black] at (9.60,-0.35)
  {$(\mu_1^E,\sigma_1^E)$};
\node[font=\small] at (11.45,-0.48) {matched base noise};
\node[smallnode] (fm) at (12.42,0.78) {continue with $f$};
\node[smallnode] (fe) at (12.42,-0.08) {continue with $f$};
\draw[model] (am.east) -- (fm.west);
\draw[model] (ae.east) -- (fe.west);
\node[align=center, font=\normalsize] at (10.70,-1.22)
  {After the first update, scorer $f$\\controls both branches.};

\end{tikzpicture}}
  \caption{CEM feedback, crossed rescoring, and the one-update splice. Left:
  a model's elite set fits the next proposal, changing the source of later
  candidates. Centre: each row holds an executed pool fixed across scorers;
  each column holds a scorer fixed across pool sources. The diagonal is the
  usual model-driven search. Right: the splice replaces only the first elite
  set, then matches the scorer and base random samples in both branches.
  The crossed comparison measures disagreement on existing pools; the splice
  changes which pools are sampled next.}
  \label{fig:crossed-design}
\end{figure}
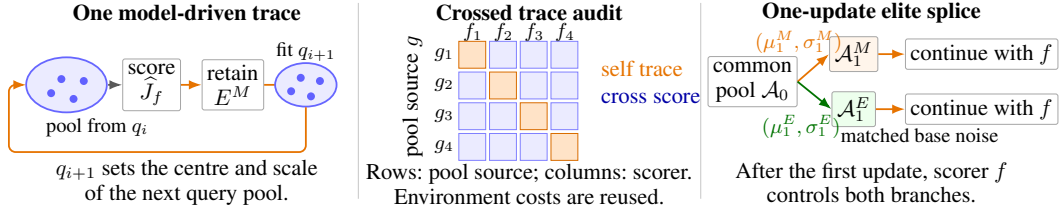

The intervention tests a consequence that rescoring alone cannot establish.
At the first update, we fit one proposal from model-ranked elites and another
from environment-ranked elites in the same pool. Both branches then continue
with the original scorer and matched random samples. This \emph{one-update
splice} changes how later candidates are generated through one earlier
decision. Its endpoint is the realised cost of the sequence selected after
the remaining CEM iterations. Figure~\ref{fig:crossed-design} shows the audit
and intervention together.

\section{Related work}

Model evaluation has long depended on the decisions that use the model.
Value-aware model learning weights errors through downstream value
functions~\citep{farahmand2017vaml}. DeepMDP connects reward and latent
transition prediction to the properties of the learned
representation~\citep{gelada2019deepmdp}. Value equivalence compares models
through their effects on specified value functions and
policies~\citep{grimm2020value}. \citet{LA+20} examine the mismatch between
one-step prediction likelihood and downstream control performance. These
works give distinct reasons to evaluate a model through its intended use.
We hold the predictors fixed and examine the finite candidate decisions
made by CEM.

\citet{yu2026evaluation} make a broader case for evaluating world models by
interventional action fidelity, policy ranking, and optimisation utility.
Their proposed reset-matched action branches are directly relevant to our
environment reference: every candidate starts from the same physical state.
\citet{bhamidipaty2026exploitable} define model exploitation through a reversal
of policy preference between the model and the environment, and analyse when
such reversals can occur. Our elite and ranking measurements examine finite
candidate disagreements; they do not estimate the worst-case exploitability
of a model over a policy class.

The closest CEM diagnostic is the CEM-stage Spearman analysis of
\citet{wang2026alignment}. It measures agreement between model and realised
cost rankings as search moves from initial candidates towards concentrated
and elite populations. This already evaluates the model on queries made
during search. Our crossed design compares a scorer's own pools with pools
generated by other surrogates. Holding the scorer fixed tests how its
disagreement varies with pool source. Holding the pool fixed compares
scorers without changing candidates. We also compare the means and widths
fitted from the two elite sets, because these parameters determine the next
queries. A rank correlation alone does not specify that refit.

\citet{li2026adjacent} execute all candidates in saved proposal pools to
distinguish failed selection from the absence of a feasible candidate.
Their adjacent-set method then reconstructs an action sequence from a
neighbourhood of low-cost candidates. We use the same basic control of
executing candidates from a common initial state. Our intervention changes
an earlier elite refit and follows its consequence through later pools. This
tests how the proposal history changes the candidates available at final
selection.

Several recent methods change components that our audit keeps fixed.
\citet{chahe2026monotone} train a cost-ranking objective that penalises
increasing action perturbations. \citet{huang2026trajlewm} add a
trajectory-level cost to endpoint scoring, and \citet{singh2026objective}
change the planning objective without retraining the predictor.
\citet{sommer2026reinforced} learn a plan-improvement operator through
imagined rollouts. These methods change candidate scores or proposal updates.
Applying the crossed comparison before and after such a change would test
which measured disagreements it reduces.

\section{Auditing a CEM update}
\label{sec:measurements}

The audit begins with a fixed planning problem: one saved simulator state and
its matching Dreamer context. We run CEM from that context, save each sampled
pool, and compare the decisions obtained from model predictions with those
obtained from executing the same candidates. Restoring the simulator state
before every execution prevents one candidate from changing the starting
point of another. Realised cost supplies the reference ordering within that
pool.

\paragraph{Candidates and elites.}
The proposal is a diagonal Gaussian: each action coordinate has its own mean
and standard deviation, with no fitted correlations between coordinates.
Here a candidate contains 15 actions, each with six coordinates. We flatten
this $15\times6$ array by listing its entries in one vector of dimension
$d=90$. The index $r$ identifies an action coordinate at one planning step;
flattening does not discard or average any actions.

Write the current proposal mean and standard deviation as $\mu$ and
$\sigma$. A candidate $j$ is sampled as
\begin{equation}
  A_j=\operatorname{clip}(\mu+\sigma\odot\epsilon_j),
  \qquad \epsilon_j\sim\mathcal N(0,I_d),\qquad j=1,\ldots,256.
  \label{eq:candidate-sampling}
\end{equation}
The product $\odot$ scales each noise coordinate by its proposal standard
deviation. Clipping enforces the action bounds after sampling. All later
comparisons use these bounded candidates, including the environment refit.

The scorer $f$ assigns a predicted cost to each sequence. Its 32 lowest-cost
sequences form $E^f$. The 32 sequences with lowest realised cost form the
environment elite set $E^E$. Both sets contain candidates from the same pool;
the environment reference supplies their realised ordering.

Elite precision records the fraction of model elites also present in the
environment elite set:
\begin{equation}
  p=\frac{|E^f\cap E^E|}{32},\qquad
  p_{\mathrm{adj}}=\frac{p-32/256}{1-32/256}.
  \label{eq:elite-precision}
\end{equation}
Two independently selected random sets of size 32 have expected overlap
four, giving expected precision $32/256=0.125$. Subtracting that baseline and
rescaling gives zero for chance overlap and one for identical sets. Negative
values indicate less overlap than the chance baseline. This measurement
focuses on which candidates CEM retains.

Pairwise order agreement covers the rest of the pool. It counts the fraction
of candidate pairs for which the scorer preserves the environment ordering,
excluding pairs tied in realised cost. Selected-candidate regret instead
subtracts the best realised cost in the pool from the realised cost of the
scorer's top candidate. It measures the cost of selection within the
available pool. Neither measurement says what candidates the next proposal
will produce. We retain Spearman agreement, Jaccard elite overlap, and
alternative elite cuts as supporting records.

\subsection{Comparing the two proposal refits}

The elite indices alone do not determine the size of an update error. Two
disjoint elite sets could contain very similar actions and fit similar
proposals. Conversely, a small change in elite membership could move the
proposal if the changed sequences lie far apart. We therefore apply CEM's
actual refit to both elite sets.

For each coordinate, CEM fits the elite mean and standard deviation, flooring
the standard deviation at 0.05. The model elites produce
$(\mu^f,\sigma^f)$; the environment elites produce
$(\mu^E,\sigma^E)$. These are \emph{next-proposal} parameters. The symbol
$\sigma$ without a superscript belongs to the \emph{current} proposal that
generated the pool.

The mean distance $D_\mu$ compares the centres of the two refits. We first
divide each difference by the corresponding current proposal width, then
take a root mean square over the 90 coordinates:
\begin{equation}
  D_{\mu}=\sqrt{\frac{1}{d}\sum_{r=1}^{d}
  \left(\frac{\mu^f_r-\mu^E_r}{\sigma_r}\right)^2}.
  \label{eq:mean-distance}
\end{equation}
$D_\mu=1$ means a root-mean-square separation of one current proposal
standard deviation. The denominator makes the scale explicit: the same raw
mean difference occupies more of the remaining search region when CEM has
already narrowed its proposal. A rise in $D_\mu$ need not mean that the raw
action difference rises.

The width comparison uses a different convention. For each coordinate, we
compare the two fitted standard deviations by their ratio:
\begin{equation}
  D_{\log\sigma}=\sqrt{\frac{1}{d}\sum_{r=1}^{d}
  \log^2\!\left(\frac{\sigma^f_r}{\sigma^E_r}\right)}.
  \label{eq:log-width-distance}
\end{equation}
A doubling and a halving have equal log magnitude. Equal fitted widths give
zero, regardless of how narrow they are. Our pre-specified distance combines
the squared mean and log-width terms with equal weight:
\begin{equation}
  D=\sqrt{\frac{D_\mu^2+D_{\log\sigma}^2}{2}}.
  \label{eq:proposal-distance}
\end{equation}
$D=0$ exactly when the two elite refits produce the same diagonal Gaussian.
Its two terms measure different kinds of disagreement, so their trajectories
must be read separately. A fall in $D$ can come from widths becoming more
similar in ratio even as the means separate in current-proposal units.

We also compute a width difference in those same current-proposal units:
\begin{equation}
  D_{\sigma}=\sqrt{\frac{1}{d}\sum_{r=1}^{d}
  \left(\frac{\sigma^f_r-\sigma^E_r}{\sigma_r}\right)^2},
  \qquad D_W=\sqrt{D_{\mu}^2+D_{\sigma}^2}.
  \label{eq:wasserstein-distance}
\end{equation}
$D_W$ is the 2-Wasserstein distance between the two diagonal Gaussian refits
after dividing coordinates by the current proposal widths, normalised by
$\sqrt d$. For diagonal Gaussians, the squared distance is the sum of squared
mean and standard-deviation differences. Equation~\ref{eq:wasserstein-distance}
applies that expression coordinatewise. Comparing $D$ with $D_W$ checks
whether the interpretation depends on using width ratios or width
differences. Neither distance alone measures candidate ordering or realised
cost.

\paragraph{The first action.}
MPC executes only the first of CEM's 15 planned actions before replanning.
A distance averaged over the full plan can conceal disagreement in that
action. We therefore also compute $D$ using only its six coordinates.

We also compare the first actions of the scorer's lowest-cost sequence and
the environment's lowest-cost sequence. Their root-mean-square gap is
reported in raw action units and after division by the current first-action
proposal widths. The raw gap records separation on the bounded action
scale. The normalised gap records separation relative to the region CEM
is still searching. A shrinking proposal can make these quantities move
in opposite directions.

Clipping and the standard-deviation floor provide two further checks on that
interpretation. We record proposal width, floor occupancy, the clipping
rate, and how far clipping displaces the sampled pool mean. These records
distinguish contraction of the proposal from changes introduced by its
bounds.

\section{Crossed audit and one-update splice}
\label{sec:crossed}

\paragraph{Crossed audit.}

Every surrogate receives the same actor-centred proposal and random samples at
iteration zero. Later pools inherit the elite decisions made along each trace.
The generator $g$ produced a pool through those preceding decisions; the
scorer $f$ ranks that fixed, realised pool.

All generator traces use the same standard-normal samples at every iteration.
Their bounded candidates match at iteration zero because their proposal
parameters also match. Later candidates differ when generators fit different
means or standard deviations. Common samples remove independent Monte Carlo
variation from the row comparison.

Scoring every generator pool with every surrogate gives 16 distances at
iteration $i$, written $D_{gfi}$. Each row reuses one candidate pool across
four scorers. Each column follows one scorer across four generated pools. The
usual model-driven trace occupies the diagonal, where $g=f$.

Each scorer has one baseline distance $D_{f0}$ on the common iteration-zero
pool. Alongside absolute distances, we show the change
$D_{gfi}-D_{f0}$. Subtracting this baseline separates a scorer's initial
level of disagreement from the change observed after generator-specific
updates. It does not remove the dependence on the distance definition.

The matrix supports comparisons that a diagonal trace cannot provide.
Suppose one scorer produces a larger distance on pools from all four
generators. That appears as a high-distance column and associates the
disagreement with current scoring. Suppose instead that all four scorers
produce larger distances on one generator's pool. That appears as a
high-distance row and associates the disagreement with the queried actions.
The physical state, task cost, and candidate count remain fixed in both
comparisons.

A particular cell may also depart from these row and column patterns.
For cell $(g,f)$, we subtract its row and column means and restore the matrix
mean:
\begin{equation}
  \text{interaction}=\text{cell distance}-\text{row mean}
  -\text{column mean}+\text{matrix mean}.
  \label{eq:crossed-decomposition}
\end{equation}
We write this two-way residual as $I_{gfi}$. The matrix mean corrects for
subtracting the overall level twice. On the diagonal, $g=f$: the same
surrogate generated and scored the pool. A positive $I_{ffi}$ means that its
own-pool distance exceeds the value predicted by adding its row and column
differences to the matrix mean. It describes a pairing discrepancy beyond
those two averages.

We summarise row, column, and residual variation by their shares of the
matrix's sum of squares. Each share is calculated within one task-seed
matrix, then averaged over the six seeds for that task. Decomposing an
already averaged matrix would discard between-seed differences in its
structure. These percentages describe the four fixed families; they are
not population variance estimates.
\ifarxivappendix
Appendix~\ref{app:attribution} gives the exact calculation.
\fi

Row differences alone do not show that an earlier elite decision caused
harm. Rescoring measures disagreement on the inherited pool; an
intervention is needed to test an update's consequence.
The distance matrix also does not measure the lowest cost available in a
pool. Even perfect agreement with the environment elites can occur when
every candidate has high realised cost. We retain realised candidate costs
to distinguish that coverage problem from disagreement within the pool.

At iteration zero, the four generator labels must point to identical
candidate IDs and environment costs. We verify this before analysing later
pools. Row differences can appear only after the first model-ranked update.

\paragraph{One-update splice.}

To test an elite decision's consequence, we fork the search at iteration
zero. Every generator has the same proposal and candidates at this point.
Using this common starting pool allows the two branches to differ in one
elite set without inheriting an earlier difference in search history.
Branch $M$ fits its next proposal from model elites; branch $E$ fits from
environment elites. Both use the same mean and standard-deviation refit,
including the width floor.

From iteration one, the original surrogate scores both branches. Matched
standard-normal samples are scaled and shifted by the two inherited proposal
states. All subsequent elite sets are selected using the original surrogate.
We reuse the model branch from the audit and execute the new candidates in
branch $E$.

After each iteration, the controller retains the lowest predicted-cost
sequence seen so far, including candidates from earlier pools. This
\emph{incumbent} is the sequence that supplies its first action if planning
stops at that point. We evaluate each branch's incumbent after iterations
one, two, and three. The primary endpoint is the branch-$E$ realised cost
minus the branch-$M$ realised cost after iteration three, the final CEM
iteration. A negative contrast favours the environment-elite branch.
The intermediate endpoints show whether the contrast changes direction
during the remaining search; they cannot replace the fixed final endpoint.

We also record the lowest realised candidate cost reached by each branch.
A branch can contain a lower-cost candidate without selecting it, so we
compare the best available cost with the incumbent's realised cost.

Only the first update is replaced; later-stage interventions are not tested.
Environment elites require realised costs for every candidate, so the splice
is an oracle diagnostic rather than a deployable planner. Its endpoint
measures a selected sequence's finite-horizon cost, not closed-loop episode
return.

\section{Experimental setting}

The crossed audit contains 12 \emph{task-seed units}: six Walker seeds and six
Cheetah seeds. Each unit has an independently trained Dreamer backbone and
actor~\citep{hafner2020dreamer}. We freeze both. The backbone maps observation
history to a 1280-dimensional context $c_t$; the actor supplies the initial CEM
mean. Four separately fitted surrogates propagate $c_t$ under proposed actions.

Each surrogate maps the current context into a feature space, propagates those
features for 15 action-conditioned linear transitions, and decodes every
predicted state back to Dreamer's context space. The rollout receives no
intermediate Dreamer states. Training minimises free-running decoded-context
error at horizons 1, 3, 5, 8, and 15, plus reward and continuation losses. A
separate validation split selects the checkpoint.

The four families change the feature map $\varphi$ and its decoder.
\emph{Identity} uses
$\varphi(c)=c$. \emph{Learned nonlinear} uses a trainable multilayer
perceptron. \emph{Anchored nonlinear} concatenates trainable affine and
nonlinear features. \emph{Random nonlinear} uses fixed random nonlinear
features. The three compressed families have width 256 and share the decoder
architecture, auxiliary heads, optimiser, data, and training budget. Identity
retains the 1280-dimensional context and is not width matched. All
four use the same action-conditioned transition form.

We use \texttt{walker\_walk} and \texttt{cheetah\_run} from the DeepMind
Control Suite~\citep{tassa2018deepmind,tunyasuvunakool2020,todorov2012mujoco}.
Seeds 0--2 form the original six-unit cohort used to decide which splice to
run. The follow-up rule required a pattern in at least two of three seeds
on each task, retained after omitting any one saved state. This limited
dependence on a single seed or decision point. The original questions and
rule were fixed before inspecting their outcomes. Later analyses of distance
components, raw action gaps and clipping explain the results without
changing the selected pairing. The rule governed follow-up computation;
it was not a significance test.
\ifarxivappendix
Appendix~\ref{app:protocol} gives its exact criteria and chronology.
\fi

Seeds 3--5 were fixed before inspecting their audit outcomes to increase
the precision of the same comparisons. They use the same state-selection,
replay, model and CEM settings and cannot change the original selection.
We also fixed the selected splice and its decision rule before inspecting
these six units' splice costs. This held-out test checks whether the same
intervention lowers final selected-sequence cost on new units. It does not
compare other families as intervention targets. We report the two cohorts
separately as well as the combined task means.

Candidate scoring begins at a saved context and predicts 15 steps. A fixed
ridge readout maps decoded contexts to the task variables in the cost. Walker
tracks a fixed goal; Cheetah tracks the frozen actor's reference trajectory.
Predicted and realised costs use the same task variables, reference targets
and cost weights. Each CEM
run has four iterations, 256 candidates, and 32 elites. The proposal width
starts at 0.35 and has a floor of 0.05. Surrogate parameters stay frozen and
training-time stochasticity is disabled.
\ifarxivappendix
Appendix Table~\ref{tab:reproducibility-settings} lists the remaining model,
training, readout, and task-cost settings.
\fi

For each unit, we export 64 simulator-state/context pairs from the frozen actor.
The audit uses one state from each of four pre-selected episode bins after a
replay check. The bins spread decision points across an episode without
selecting states for their audit outcome. We restore the exact state before
every candidate execution and verify repeated costs. Each generator trace
uses the same standard-normal samples at every iteration; differences in
its proposal parameters produce differences in its later bounded actions.

The complete crossed audit executes 196,608 candidates; the original six-unit
cohort accounts for 98,304. Each pool is executed once and then scored by all four
surrogates, so the off-diagonal cells add model evaluations without adding
simulator rollouts. We average states within each task-seed unit before
comparing seeds. The diagonal trajectories also average the four model
families; crossed matrices retain every generator-scorer cell. Candidate
sequences, matrix cells, and saved states are repeated measurements within
a unit, not independent replicates.

\section{Results}

\subsection{The proposal distance falls without better candidate ranking}

\begin{figure}[t]
  \centering
  \includegraphics[width=0.98\textwidth]{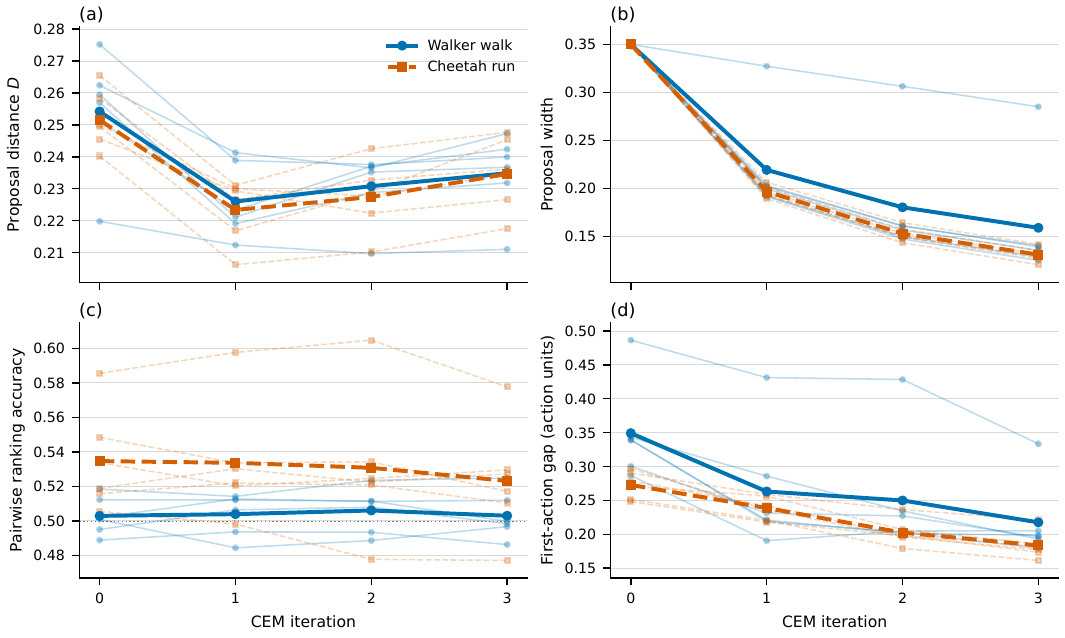}
  \caption{Diagonal audit measurements over four CEM iterations. Thin lines
  show six task-seed units per task after averaging the four surrogates; heavy
  lines show task means. The model bank, candidate count, elite cut, and horizon
  are fixed. From iteration zero to three, $D$ and the raw first-action gap
  decrease with proposal width. Pairwise order agreement stays near chance
  on Walker and declines on Cheetah.
  The dotted line in (c) marks chance agreement.}
  \label{fig:pilot-traces}
\end{figure}

The pre-specified proposal distance decreases from iteration zero to three
in all original and extension units. In the combined 12-unit result,
Walker's task mean falls from
$D=\PTWalkerDInitial$ at iteration zero to
$\PTWalkerDFinal$ at iteration three; Cheetah falls from
$\PTCheetahDInitial$ to $\PTCheetahDFinal$
(Figure~\ref{fig:pilot-traces}). Every task-seed contrast is negative.
They range from $\PTWalkerDContrastMinimum$ to
$\PTWalkerDContrastMaximum$ on Walker and from
$\PTCheetahDContrastMinimum$ to $\PTCheetahDContrastMaximum$ on Cheetah.

The two components of $D$ move in opposite directions
(Table~\ref{tab:distance-components}). The mean term $D_\mu$ rises on both
tasks, so the two elite refits separate relative to the current proposal
width. The log-width term falls. Its reduction accounts for the lower $D$
despite the rise in mean disagreement. The alternative distance $D_W$
rises by $\PTWalkerDWassersteinPercentIncrease\%$
on Walker and $\PTCheetahDWassersteinPercentIncrease\%$ on Cheetah. A falling
$D$ records a reduction in its ratio-based width term. It does not establish
that CEM retains more of the environment elites or selects a lower-cost
sequence.

The first-action gap gives the same warning about scale. The scorer's
selected action and the environment's selected action become closer in raw
action units, yet farther apart relative to the current proposal width.
CEM's search region contracts faster than their separation. This scale
dependence is already present at the first planned step.

\begin{table}[t]
  \centering
  \begin{tabular}{lcccc}
    \toprule
    & \multicolumn{2}{c}{Walker} & \multicolumn{2}{c}{Cheetah} \\
    \cmidrule(lr){2-3}\cmidrule(lr){4-5}
    Measurement & Iter. 0 & Iter. 3 & Iter. 0 & Iter. 3 \\
    \midrule
    Mean distance $D_\mu$ & \PTWalkerDMeanInitial & \PTWalkerDMeanFinal & \PTCheetahDMeanInitial & \PTCheetahDMeanFinal \\
    Log-width distance $D_{\log\sigma}$ & \PTWalkerDLogWidthInitial & \PTWalkerDLogWidthFinal & \PTCheetahDLogWidthInitial & \PTCheetahDLogWidthFinal \\
    Linear-width distance $D_\sigma$ & \PTWalkerDLinearWidthInitial & \PTWalkerDLinearWidthFinal & \PTCheetahDLinearWidthInitial & \PTCheetahDLinearWidthFinal \\
    Whitened distance $D_W$ & \PTWalkerDWassersteinInitial & \PTWalkerDWassersteinFinal & \PTCheetahDWassersteinInitial & \PTCheetahDWassersteinFinal \\
    First-action proposal distance $D$ & \PTWalkerFirstActionDInitial & \PTWalkerFirstActionDFinal & \PTCheetahFirstActionDInitial & \PTCheetahFirstActionDFinal \\
    First-action gap (action units) & \PTWalkerFirstActionGapRawInitial & \PTWalkerFirstActionGapRawFinal & \PTCheetahFirstActionGapRawInitial & \PTCheetahFirstActionGapRawFinal \\
    First-action gap (proposal SD) & \PTWalkerFirstActionGapNormalisedInitial & \PTWalkerFirstActionGapNormalisedFinal & \PTCheetahFirstActionGapNormalisedInitial & \PTCheetahFirstActionGapNormalisedFinal \\
    \bottomrule
  \end{tabular}
  \caption{Components of the proposal refit and first-action disagreement.
  Values average diagonal model traces and states within task-seed units,
  then the six seeds per task. The candidate count, elite cut, and horizon
  remain fixed. Fitted means separate in current-proposal units as width
  disagreement falls. Raw first-action gaps shrink, but width-normalised
  gaps grow. These quantities cannot be read as interchangeable measures of
  elite agreement.}
  \label{tab:distance-components}
\end{table}

At iteration three,
only $\PTWalkerFloorOccupancyFinalPercent\%$ of Walker coordinates and
$\PTCheetahFloorOccupancyFinalPercent\%$ of Cheetah coordinates are at the
width floor. Most widths therefore contract without reaching the minimum.
The mean displacement from clipping also decreases.
\ifarxivappendix
Table~\ref{tab:app-bound-effects} gives its endpoints, and
Figure~\ref{fig:app-clipping} records clipping rates along the traces.
\fi

Pairwise order agreement changes from
$\PTWalkerPairwiseAccuracyInitial$ to $\PTWalkerPairwiseAccuracyFinal$ on
Walker and from $\PTCheetahPairwiseAccuracyInitial$ to
$\PTCheetahPairwiseAccuracyFinal$ on Cheetah. Chance-adjusted elite precision
changes from $\PTWalkerAdjustedElitePrecisionInitial$ to
$\PTWalkerAdjustedElitePrecisionFinal$ on Walker and from
$\PTCheetahAdjustedElitePrecisionInitial$ to
$\PTCheetahAdjustedElitePrecisionFinal$ on Cheetah. On the original seeds
0--2, the neighbouring elite cuts keep task-mean adjusted precision below 0.10.
\ifarxivappendix
Figure~\ref{fig:app-elite-cuts} reports these cuts.
\fi

\subsection{Scorer variation dominates on Cheetah}

\begin{figure}[t]
  \centering
  \includegraphics[width=0.94\textwidth]{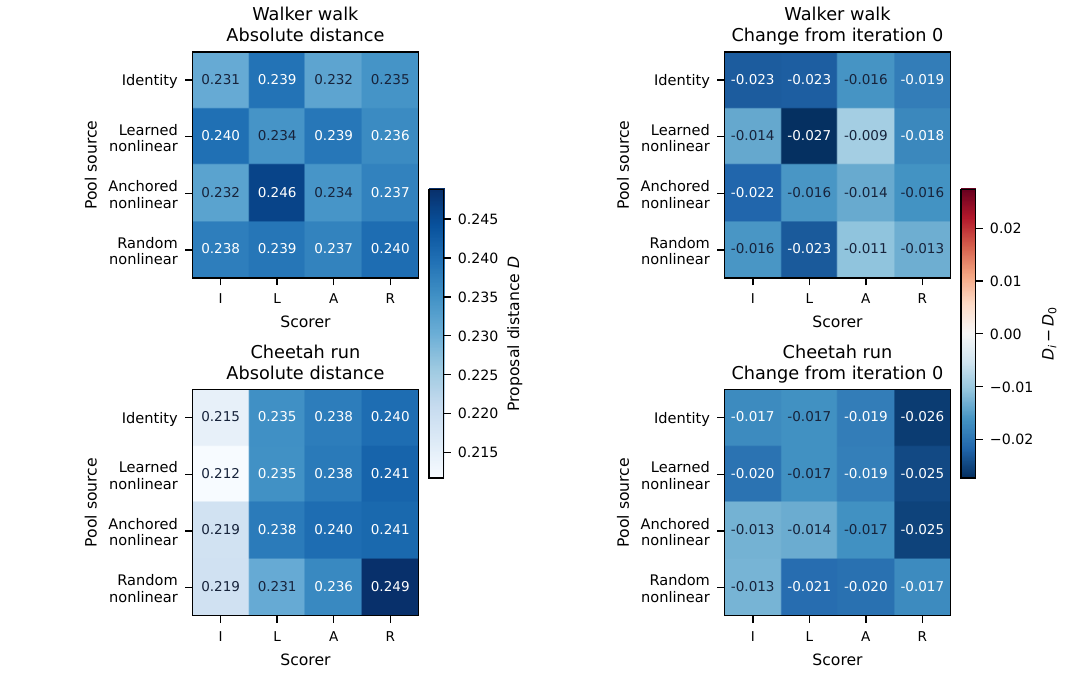}
  \caption{Final-iteration crossed proposal distance, averaged over six seeds
  per task. The attribution percentages instead average the six within-seed
  decompositions described in Section~\ref{sec:crossed}. A row fixes the
  surrogate that generated
  a pool; each column rescores that executed pool with Identity, Learned
  nonlinear, Anchored nonlinear, or Random nonlinear. The right panels
  subtract each scorer's value on the common initial pool. Cheetah has greater
  variation between scorers than between pool sources in all six seeds.
  Walker divides variation across pool source, scorer, and pairing.}
  \label{fig:pilot-matrices}
\end{figure}

At iteration three, scorer columns account for
$\PTCheetahAttributionScorerPercent\%$ of the descriptive sum of squares on
Cheetah; pool-source rows account for $\PTCheetahAttributionPoolPercent\%$ and
pairings for $\PTCheetahAttributionPairingPercent\%$. Walker is mixed: the
corresponding fractions are $\PTWalkerAttributionScorerPercent\%$,
$\PTWalkerAttributionPoolPercent\%$, and
$\PTWalkerAttributionPairingPercent\%$ (Figure~\ref{fig:pilot-matrices}).
Scorer variation exceeds pool-source variation in
$\PTCheetahAttributionScorerExceedsPoolSeedCount$ of
$\PTCheetahAttributionSeedCount$ Cheetah seeds, compared with
$\PTWalkerAttributionScorerExceedsPoolSeedCount$ of
$\PTWalkerAttributionSeedCount$ Walker seeds. Thus the Cheetah column
structure is present within individual seeds, rather than appearing only
after their matrices are averaged.

Random nonlinear has a positive own-pool residual in all six original units.
Its distance on its own generated pool exceeds the additive value from
its row and column averages. Every one-state omission preserves the sign.
\ifarxivappendix
Appendix~\ref{app:protocol} gives the seed-level residuals, and
Table~\ref{tab:app-loo-attribution} gives the omitted-state results.
\fi
In the combined audit, the residual is positive in
\PTRandomNonlinearWalkerPositiveResidualCountFinal{} of
\PTRandomNonlinearWalkerResidualSeedCount{} Walker seeds and
\PTRandomNonlinearCheetahPositiveResidualCountFinal{} of
\PTRandomNonlinearCheetahResidualSeedCount{} Cheetah seeds. The original
decision rule therefore selected Random nonlinear for the splice. No
pool-source row met the cross-task criterion.

\subsection{Replacing the first elite set lowers final selected-sequence cost}

The pre-specified Random nonlinear splice replaces the first model-elite
update with the environment-elite update and then returns both branches to the
same surrogate. Before averaging within a task-seed unit, we divide each
state-level cost contrast by the difference between the 90th and 10th
percentiles of realised cost in its shared initial pool. This is the pool's
\emph{interdecile range}. A value of $-0.5$, for example, means that the
environment-elite branch selects a sequence whose realised cost is lower by
half of that initial-pool range.

The final contrast is negative in all \PTSpliceSelectionFinalUnitCount{}
selection units. The task means are
\PTSpliceSelectionWalkerFinalNormalisedMean{} on Walker and
\PTSpliceSelectionCheetahFinalNormalisedMean{} on Cheetah.

\begin{figure}[t]
  \centering
  \includegraphics[width=0.90\textwidth]{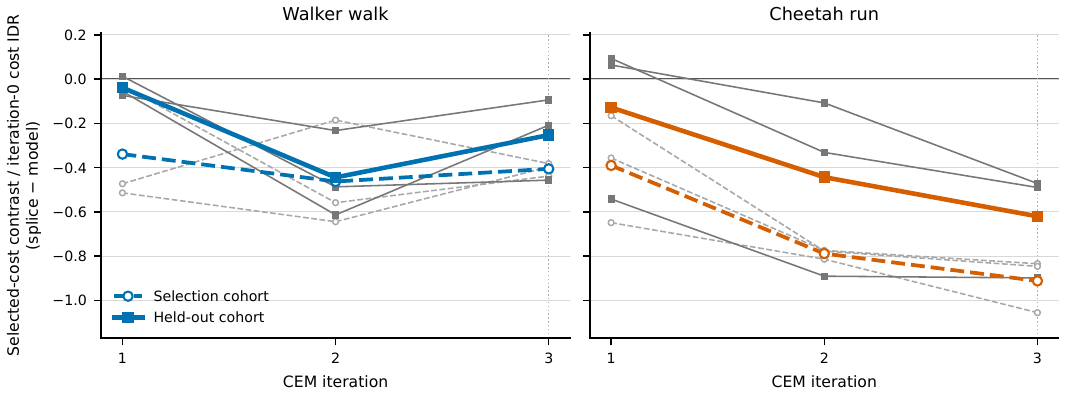}
  \caption{Effect of replacing the first Random nonlinear elite set on selected
  cost over iterations one to three. Both branches share the initial pool,
  later scorer, and base random samples; only the first elite set differs.
  Environment-elite minus model contrasts are scaled by the initial pool's
  realised-cost interdecile range. Dashed circles show selection units, solid
  squares show held-out units, and lines show cohort means. Every final
  contrast is negative, including all six held-out units; intermediate
  contrasts can be positive.
  \ifarxivappendix Table~\ref{tab:elite-splice-primary} reports all 12
  task-seed trajectories.\fi}
  \label{fig:elite-splice}
\end{figure}

The direction repeats in all \PTSpliceHeldoutFinalUnitCount{} held-out
units. Their task means are \PTSpliceHeldoutWalkerFinalNormalisedMean{} on
Walker and \PTSpliceHeldoutCheetahFinalNormalisedMean{} on Cheetah. The fixed
cross-task rule survives all \PTSpliceHeldoutStateOmissionCount{} one-state
omissions. At least two of three seeds per task retain a negative contrast after each
omission. Omitting one state makes Walker seed 5 slightly positive.
\ifarxivappendix
Table~\ref{tab:elite-splice-heldout-loso} records the exception.
\fi
Across both cohorts, all \PTSpliceCombinedFinalUnitCount{}
task-seed contrasts are negative; the combined descriptive means are
\PTSpliceCombinedWalkerFinalNormalisedMean{} and
\PTSpliceCombinedCheetahFinalNormalisedMean{} (Figure~\ref{fig:elite-splice}).

The best realised candidate also has lower cost in every task-seed unit. Its
combined contrast is \PTSpliceCombinedWalkerFinalBestNormalisedMean{} initial-pool
ranges on Walker and \PTSpliceCombinedCheetahFinalBestNormalisedMean{} on
Cheetah. The Cheetah splice therefore reaches pools containing lower-cost
sequences. Walker's best-candidate change is smaller than its selected-cost
change, indicating that coverage alone does not describe the selected-cost
contrast. Later model selection also differs between the branches. Several
held-out units have a positive contrast after iteration one. Replacing the
first elite set does not improve every intermediate endpoint; its benefit
appears at the fixed final endpoint.

\section{Discussion and limitations}

A proposal audit needs an explicit reference scale. Mean separation in raw
action units answers a different question from separation relative to the
remaining search width. Width ratios answer another. Reporting those
components alongside elite recovery prevents contraction alone from being
read as improved decision agreement. Realised selected-sequence cost is
still needed to test whether an update changes the outcome of search.

The crossed comparison guides the next controlled test. On Cheetah, a scoring
change can be tested across several fixed pool sources, because the observed
differences follow scorer columns in every seed. On Walker, a score-only
account omits row and pairing variation. A proposal constraint could be
tested with the scorer held fixed, or a scoring change could be tested on
the same saved pools. The matrix makes these comparisons available without
additional off-diagonal simulator executions. It does not choose a remedy
or guarantee that one will work.

The Random nonlinear splice shows that one elite decision can affect the
selected cost after three further CEM iterations. Its direction
repeats on units held out from pairing selection. Other surrogates might
also benefit from an environment-elite update: we have not compared splices
across all four families. The result therefore establishes the consequence
of the selected update, without establishing that the residual identified
the best intervention target.

The experiment covers two tasks, six independently trained Dreamer
backbones per task, four states per backbone, one horizon, and one CEM
configuration. The shared transition form controls part of the model
comparison; other dynamics classes may produce different matrices.
Identity also retains a wider context than the three compressed families,
so family differences cannot be assigned to feature type alone. The four
states sample a restricted part of each actor's experience, and state
omissions check dependence on those points rather than coverage of all
reachable states.

The audit requires a resettable environment and execution of every
candidate. It is suited to diagnosing a model-planner loop in simulation.
The splice uses information unavailable to an ordinary deployed planner,
and lower finite-horizon selected-sequence cost does not establish higher
closed-loop return. A method motivated by the audit must be evaluated
without that oracle and under receding-horizon control.

\paragraph{Conclusion.}
A world model used by CEM determines both the ordering of current candidates
and the distribution of later queries. Crossed rescoring describes these
roles separately; a matched one-update splice tests whether an elite
decision changes the final selected-sequence cost.

\section*{Acknowledgements}
This research includes computations using the computational cluster
Katana~\citep{unsw2010katana} supported by Research Technology Services
at UNSW Sydney.

\ifarxivappendix
\clearpage
\appendix

\section{Follow-up rule and experiment chronology}
\label{app:protocol}

The original audit used Walker and Cheetah seeds 0--2, with four saved
states, four surrogate families, and four CEM iterations per task-seed
unit. Its gate was a rule fixed before outcome inspection for deciding
whether to extend the audit and run an intervention. It was not a significance test.

\paragraph{Original questions.}
The first question tested whether the family-mean diagonal distance $D$
increased from iteration zero to three. The second tested whether pool
source or an own-pool pairing produced a replicated pattern at iteration
three. A pool-source pattern required at least three of four scorer columns
to name the same highest-distance generator within a task; that generator
could differ between tasks. A pairing pattern required the same surrogate
to have a positive diagonal residual $I_{ff3}$ on both tasks.

A qualifying pattern needed the stated direction on Walker and Cheetah,
at least two of three seeds per task, and the same outcome after each of
the four one-state omissions. The omissions recomputed task-seed summaries
from the remaining three states. This rule limited reliance on one seed or
one saved state. It did not estimate the probability of replication on a
new task or provide a multiple-testing correction. Only the second
question could trigger the one-update splice.

The original measurements included $D$, pairwise order agreement, and
elite recovery. Selected-candidate regret, alternative elite cuts, and
downstream family ordering did not enter the gate. We specified the
change-from-baseline matrices, descriptive sums of squares, random-elite
reference and $D_W$ after inspecting the Walker results, before inspecting
the completed Cheetah summaries. The mean/width decomposition, raw-action
gap, floor occupancy and clipping displacement followed inspection of the
six-unit pilot and preceded completion of the extension cohort. These
analyses could not change the original rule or the selected family.

\paragraph{Gate outcome.}
$D$ decreased, contrary to the original increasing-distance hypothesis.
No pool-source pattern qualified under the seed
and state-omission checks. Random nonlinear's positive own-pool residual
qualified and selected its splice. At iteration three, $I_{ff3}$ is
$\PTRandomNonlinearWalkerSeedZeroResidualFinal$,
$\PTRandomNonlinearWalkerSeedOneResidualFinal$, and
$\PTRandomNonlinearWalkerSeedTwoResidualFinal$ in Walker seeds 0, 1, and 2;
the Cheetah values are
$\PTRandomNonlinearCheetahSeedZeroResidualFinal$,
$\PTRandomNonlinearCheetahSeedOneResidualFinal$, and
$\PTRandomNonlinearCheetahSeedTwoResidualFinal$. Every one-state omission
preserves their positive sign
(Table~\ref{tab:app-loo-attribution}). No splice cost was used for this
selection.

\paragraph{Precision extension and held-out splice.}
Seeds 3--5 were fixed as an audit precision extension before their audit
outcomes were inspected. They use the same model, state-selection, replay,
and CEM settings. They do not reopen the original gate. After Random
nonlinear was selected on seeds 0--2, the same first-update splice and its
held-out decision rule were fixed before inspecting seeds 3--5 splice
costs. The rule requires a negative final selected-cost contrast in at
least two of three seeds on each task, including after every one-state
omission. All six full-state contrasts are negative. The rule also passes
under every omission, although omitting bin 3 makes Walker seed 5 slightly
positive (Table~\ref{tab:elite-splice-heldout-loso}).

The first update is the intervention point because all traces then share
the same pool. Iteration three is the primary endpoint because it ends
the fixed four-iteration search. Intermediate contrasts are reported at
iterations one and two, including positive contrasts. We did not select
an intervention stage or endpoint by its realised gain. This experiment
tests the selected update's consequence and its repetition on held-out
units; it does not compare the four families as intervention targets.

\section{Descriptive attribution calculation}
\label{app:attribution}

At iteration $i$, the state-averaged task-seed matrix contains $K=4$
generator rows and $K=4$ scorer columns. Write its row, column, and grand
means as $\bar D_{g\cdot i}$, $\bar D_{\cdot f i}$, and
$\bar D_{\cdot\cdot i}$. Define the row and column deviations from the
grand mean by
\begin{equation}
  R_{gi}=\bar D_{g\cdot i}-\bar D_{\cdot\cdot i},\qquad
  C_{fi}=\bar D_{\cdot f i}-\bar D_{\cdot\cdot i}.
\end{equation}
The remaining cell residual is
$I_{gfi}=D_{gfi}-\bar D_{\cdot\cdot i}-R_{gi}-C_{fi}$, equivalent to
Equation~\ref{eq:crossed-decomposition}. The sums of squares are
\begin{align}
  {\rm SS}_{\rm row}&=K\sum_g R_{gi}^2,&
  {\rm SS}_{\rm col}&=K\sum_f C_{fi}^2,\\
  {\rm SS}_{\rm res}&=\sum_{g,f}I_{gfi}^2,&
  {\rm SS}_{\rm tot}&=\sum_{g,f}
       (D_{gfi}-\bar D_{\cdot\cdot i})^2.
\end{align}
They satisfy
${\rm SS}_{\rm tot}={\rm SS}_{\rm row}+{\rm SS}_{\rm col}+{\rm SS}_{\rm res}$.
For each component $x$, we compute
$100\,{\rm SS}_x/{\rm SS}_{\rm tot}$ within each task-seed matrix, then
average the six shares for a task. We do not decompose the six-seed mean
matrix to obtain the reported percentages. The families are fixed test
models, so the shares describe this matrix rather than population
variance components or causal effects.

\section{Seed-level crossed matrices}
\label{app:seed-matrices}

\begin{table}[H]
  \centering
  \small
  \setlength{\tabcolsep}{7pt}
\begin{tabular}{lcccc}
  \toprule
  & \multicolumn{2}{c}{Walker} & \multicolumn{2}{c}{Cheetah} \\
  \cmidrule(lr){2-3}\cmidrule(lr){4-5}
  Metric & Iter. 0 & Iter. 3 & Iter. 0 & Iter. 3 \\
  \midrule
  Proposal distance $D$ & \PTWalkerDInitial & \PTWalkerDFinal & \PTCheetahDInitial & \PTCheetahDFinal \\
  Proposal width & \PTWalkerProposalWidthInitial & \PTWalkerProposalWidthFinal & \PTCheetahProposalWidthInitial & \PTCheetahProposalWidthFinal \\
  Whitened distance $D_W$ & \PTWalkerDWassersteinInitial & \PTWalkerDWassersteinFinal & \PTCheetahDWassersteinInitial & \PTCheetahDWassersteinFinal \\
  First-action gap (action units) & \PTWalkerFirstActionGapRawInitial & \PTWalkerFirstActionGapRawFinal & \PTCheetahFirstActionGapRawInitial & \PTCheetahFirstActionGapRawFinal \\
  First-action gap (proposal SD) & \PTWalkerFirstActionGapNormalisedInitial & \PTWalkerFirstActionGapNormalisedFinal & \PTCheetahFirstActionGapNormalisedInitial & \PTCheetahFirstActionGapNormalisedFinal \\
  Pairwise order agreement & \PTWalkerPairwiseAccuracyInitial & \PTWalkerPairwiseAccuracyFinal & \PTCheetahPairwiseAccuracyInitial & \PTCheetahPairwiseAccuracyFinal \\
  Adjusted elite precision & $\PTWalkerAdjustedElitePrecisionInitial$ & $\PTWalkerAdjustedElitePrecisionFinal$ & \PTCheetahAdjustedElitePrecisionInitial & \PTCheetahAdjustedElitePrecisionFinal \\
  \bottomrule
\end{tabular}

  \caption{Task means at the common initial pool and final CEM iteration. The
  horizon, candidate count, elite cut, and four-surrogate average are fixed.
  $D$ and the raw first-action gap decrease; $D_W$ and the width-normalised
  action gap increase. Pairwise order agreement stays near chance on Walker
  and declines on Cheetah. Chance-adjusted elite precision does not improve.}
  \label{tab:pilot-diagnostics}
\end{table}

Figures~\ref{fig:app-seed-matrices-absolute}
and~\ref{fig:app-seed-matrices-change} retain each of the six original gate
units before task averaging. The main figures add the six extension
units. Rows and columns follow Identity, Learned nonlinear, Anchored nonlinear,
and Random nonlinear; the axes use the first word of each name.

\begin{figure}[H]
  \centering
  \includegraphics[width=0.98\textwidth]{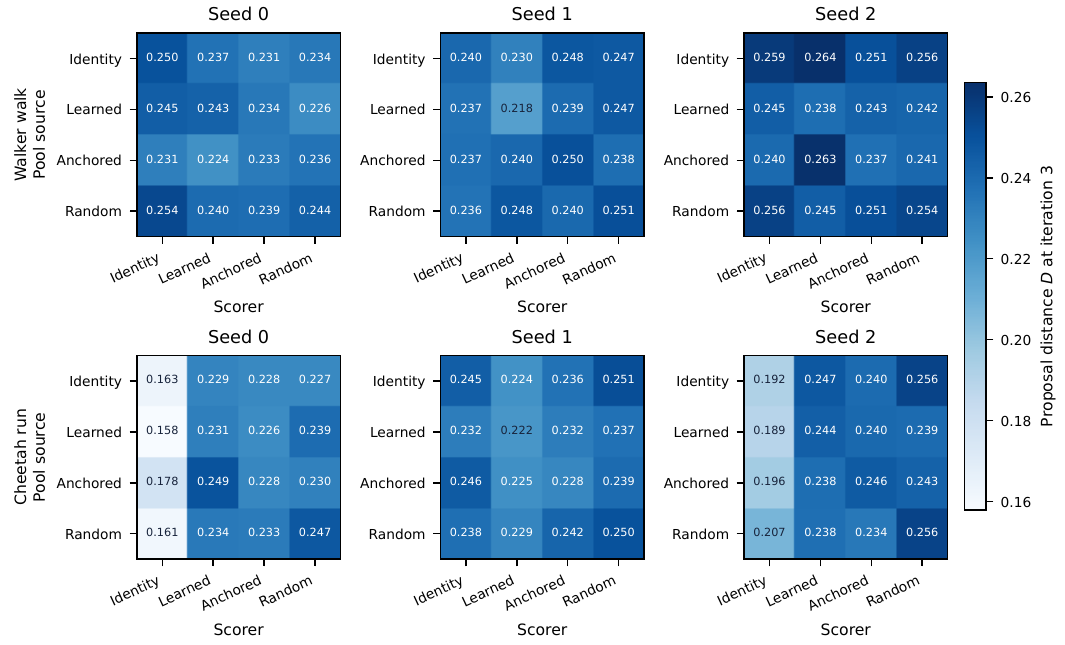}
  \caption{Final-iteration crossed proposal distance for the six seed 0--2
  gate units. A row fixes the generator pool and columns rescore its executed
  candidates; candidate sets are never pooled across seeds. Cheetah is
  scorer-dominated in all three gate units. Identity has the lowest column
  mean in seeds 0 and 2;
  Learned nonlinear has the lowest in seed 1.}
  \label{fig:app-seed-matrices-absolute}
\end{figure}

\begin{figure}[H]
  \centering
  \includegraphics[width=0.98\textwidth]{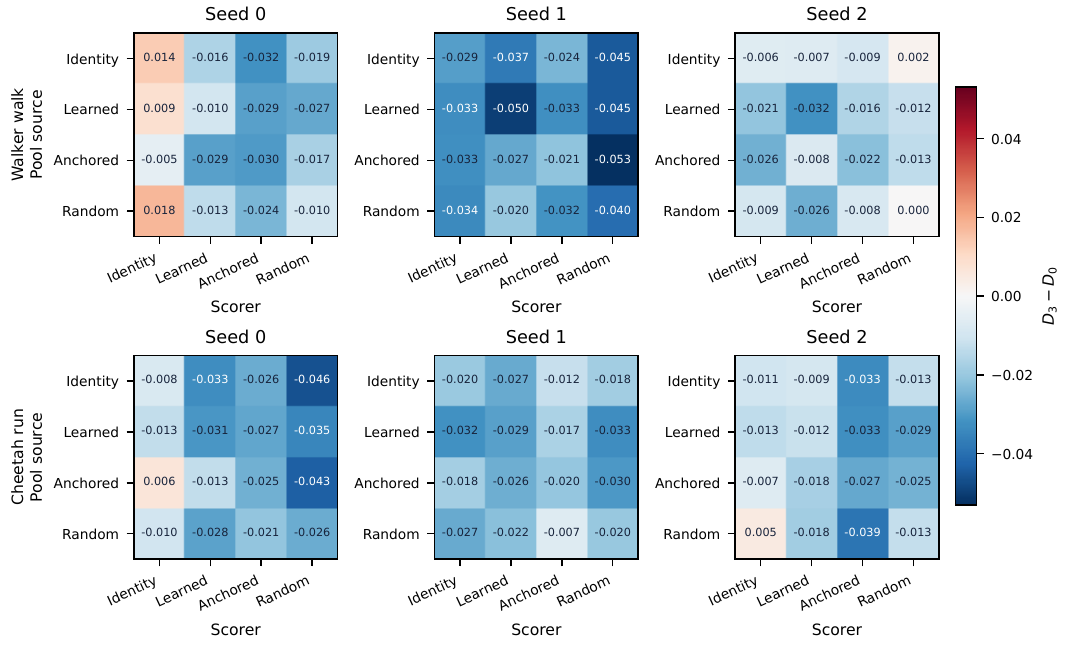}
  \caption{Change in crossed proposal distance for the six gate units. Each
  cell holds its scorer fixed and subtracts that scorer's value on the common
  iteration-zero pool. Most cells are negative at iteration three. This
  comparison records the ratio-based distance $D$ and does not determine
  candidate order agreement.}
  \label{fig:app-seed-matrices-change}
\end{figure}

\section{Leave-one-state-out checks}
\label{app:state-omission}

Each task--seed unit contains four saved states drawn from fixed replay time
bins. We repeat the gate and crossed attribution after omitting one bin at a
time. We average the remaining three states within each task--seed unit
before comparing seeds.

\begin{table}[H]
  \centering
  \footnotesize
  \resizebox{0.96\textwidth}{!}{%
\begin{tabular}{llrrrrc}
\toprule
Omitted bin & Task & Seed 0 & Seed 1 & Seed 2 & $n_+$ & Pass \\
\midrule
All states & Walker walk & -0.009 & -0.035 & -0.015 & 0 & no \\
All states & Cheetah run & -0.023 & -0.022 & -0.016 & 0 & no \\
\addlinespace
$-1$ & Walker walk & -0.011 & -0.027 & -0.011 & 0 & no \\
$-1$ & Cheetah run & -0.030 & -0.025 & -0.019 & 0 & no \\
\addlinespace
$-3$ & Walker walk & -0.000 & -0.040 & -0.020 & 0 & no \\
$-3$ & Cheetah run & -0.020 & -0.017 & -0.025 & 0 & no \\
\addlinespace
$-5$ & Walker walk & -0.018 & -0.035 & -0.021 & 0 & no \\
$-5$ & Cheetah run & -0.018 & -0.028 & -0.011 & 0 & no \\
\addlinespace
$-7$ & Walker walk & -0.007 & -0.039 & -0.008 & 0 & no \\
$-7$ & Cheetah run & -0.023 & -0.019 & -0.007 & 0 & no \\
\bottomrule
\end{tabular}
  }
  \caption{Leave-one-state-out check for the proposal-distance gate. Each
  column is a task--seed contrast after averaging model families; $n_+$ counts
  contrasts above zero. The task, seeds, CEM settings, and remaining three
  states are fixed within an omission. Every omission preserves a negative
  iteration-three minus iteration-zero contrast on both tasks.}
  \label{tab:app-loo-gate}
\end{table}

\begin{table}[H]
  \centering
  \scriptsize
  \resizebox{\textwidth}{!}{%
\begin{tabular}{llllcc}
\toprule
Omitted bin & Walker support & Cheetah support & Cross-task self interaction & Support & Interaction \\
\midrule
All states & none & none & Identity; Random nonlinear & no & yes \\
$-1$ & none & none & Random nonlinear & no & yes \\
$-3$ & none & none & Identity; Random nonlinear & no & yes \\
$-5$ & none & none & Identity; Random nonlinear & no & yes \\
$-7$ & Random nonlinear & none & Random nonlinear & no & yes \\
\bottomrule
\end{tabular}
  }
  \caption{Leave-one-state-out check for the two attribution triggers. A
  support trigger requires at least three scorer columns to select the same
  highest-distance generator within each task. An interaction trigger requires
  the same surrogate's own-pool residual to be positive on both tasks. Random
  nonlinear passes under all four omissions; no generator passes the support
  trigger.}
  \label{tab:app-loo-attribution}
\end{table}

\section{Seed-level elite-splice results}
\label{app:elite-splice}

The main paper plots the paired normalised contrasts.
Table~\ref{tab:elite-splice-primary} retains the fixed raw endpoint and
every task--seed trajectory for the selection and held-out cohorts.
The environment-elite branch differs only in the first update; iterations one
to three again use Random nonlinear scores in both branches.

\begin{table}[H]
  \centering
  \scriptsize
  \resizebox{\textwidth}{!}{%
\begin{tabular}{lllrrrrr}
\toprule
Cohort & Task & Seed & $\Delta_1/s_0$ & $\Delta_2/s_0$ & $\Delta_3$ & $\Delta_3/s_0$ & $\Delta^{\mathrm{best}}_3/s_0$ \\
\midrule
Selection & Walker walk & 0 & $-0.029$ & $-0.558$ & $-1409.812$ & \textbf{$-0.438$} & $-0.018$ \\
 &  & 1 & $-0.514$ & $-0.644$ & $-1304.868$ & \textbf{$-0.396$} & $-0.052$ \\
 &  & 2 & $-0.473$ & $-0.185$ & $-801.631$ & \textbf{$-0.381$} & $-0.049$ \\
\addlinespace
 & Cheetah run & 0 & $-0.356$ & $-0.774$ & $-3124.093$ & \textbf{$-0.834$} & $-0.379$ \\
 &  & 1 & $-0.649$ & $-0.814$ & $-2880.493$ & \textbf{$-1.055$} & $-0.431$ \\
 &  & 2 & $-0.165$ & $-0.777$ & $-1852.621$ & \textbf{$-0.846$} & $-0.494$ \\
\midrule
Held-out & Walker walk & 3 & $-0.055$ & $-0.613$ & $-710.564$ & \textbf{$-0.208$} & $-0.139$ \\
 &  & 4 & $+0.013$ & $-0.487$ & $-1776.227$ & \textbf{$-0.456$} & $-0.051$ \\
 &  & 5 & $-0.073$ & $-0.232$ & $-760.797$ & \textbf{$-0.093$} & $-0.008$ \\
\addlinespace
 & Cheetah run & 3 & $+0.094$ & $-0.331$ & $-1345.455$ & \textbf{$-0.489$} & $-0.528$ \\
 &  & 4 & $-0.542$ & $-0.890$ & $-3394.181$ & \textbf{$-0.898$} & $-0.500$ \\
 &  & 5 & $+0.064$ & $-0.107$ & $-1125.465$ & \textbf{$-0.471$} & $-0.318$ \\
\bottomrule
\end{tabular}
  }
  \caption{Paired realised-cost contrasts for the 12 Random nonlinear splice
  units. Seeds 0--2 selected the pairing without using splice costs; seeds
  3--5 test the fixed intervention without using their splice outcomes for
  selection. Both branches hold the initial pool, later scorer, and base
  samples fixed. $\Delta_i$ is environment-elite branch minus model branch
  after iteration $i$, $\Delta^{\mathrm{best}}_3$ compares the best realised
  candidates reached, and $s_0$ is the realised-cost interdecile range in the
  shared initial pool. Every iteration-three selected-cost contrast is
  negative.}
  \label{tab:elite-splice-primary}
\end{table}

\begin{table}[H]
  \centering
  \scriptsize
  \resizebox{\textwidth}{!}{%
\begin{tabular}{lrrrrrrc}
\toprule
Omitted bin & W3 & W4 & W5 & C3 & C4 & C5 & Rule \\
\midrule
None & $-0.208$ & $-0.456$ & $-0.093$ & $-0.489$ & $-0.898$ & $-0.471$ & Pass \\
1 & $-0.228$ & $-0.523$ & $-0.250$ & $-0.548$ & $-1.003$ & $-0.302$ & Pass \\
3 & $-0.170$ & $-0.381$ & $+0.001$ & $-0.478$ & $-0.651$ & $-0.484$ & Pass \\
5 & $-0.164$ & $-0.315$ & $-0.006$ & $-0.542$ & $-1.025$ & $-0.561$ & Pass \\
7 & $-0.269$ & $-0.606$ & $-0.119$ & $-0.391$ & $-0.915$ & $-0.539$ & Pass \\
\bottomrule
\end{tabular}
  }
  \caption{Leave-one-state-out check for the held-out splice. Entries
  are final normalised selected-cost contrasts for Walker (W) and Cheetah (C).
  The family, CEM settings, and remaining three states are fixed. Omitting bin
  3 makes Walker seed 5 slightly positive, but the pre-specified rule still
  passes with two negative Walker seeds and three negative Cheetah seeds. All
  other omissions retain six negative contrasts.}
  \label{tab:elite-splice-heldout-loso}
\end{table}

\begin{table}[H]
  \centering
  \footnotesize
  \setlength{\tabcolsep}{5pt}
  \begin{tabularx}{\textwidth}{p{0.25\textwidth}X}
    \toprule
    Audit pattern & Reading and next check \\
    \midrule
    Scorer columns dominate & Test a score change or ranking loss against realised pairwise order and elite recovery. \\
    Pool-source rows dominate & Test a proposal constraint or actor-anchored sampling. \\
    Positive diagonal residual & Run a targeted model--planner intervention. \\
    Metrics split as width contracts & Report width, ranking, and action-level fidelity. \\
    \bottomrule
  \end{tabularx}
  \caption{Matrix patterns and possible controlled follow-ups. The
  candidate pool is fixed across scorer columns, and the scorer is fixed down a
  column. Row, column, and diagonal patterns therefore direct the next test to
  a different component; none establishes that a proposed response will work.}
  \label{tab:diagnostic-reading}
\end{table}

\section{Elite-set size}
\label{app:elite-cuts}

The primary CEM update uses 32 elites. We recompute the final diagonal
measurements with 8, 16, and 64 elites from the saved candidate pools for
the original seeds 0--2. These are alternative refits of existing pools;
they do not generate new CEM traces.

\begin{figure}[H]
  \centering
  \includegraphics[width=0.98\textwidth]{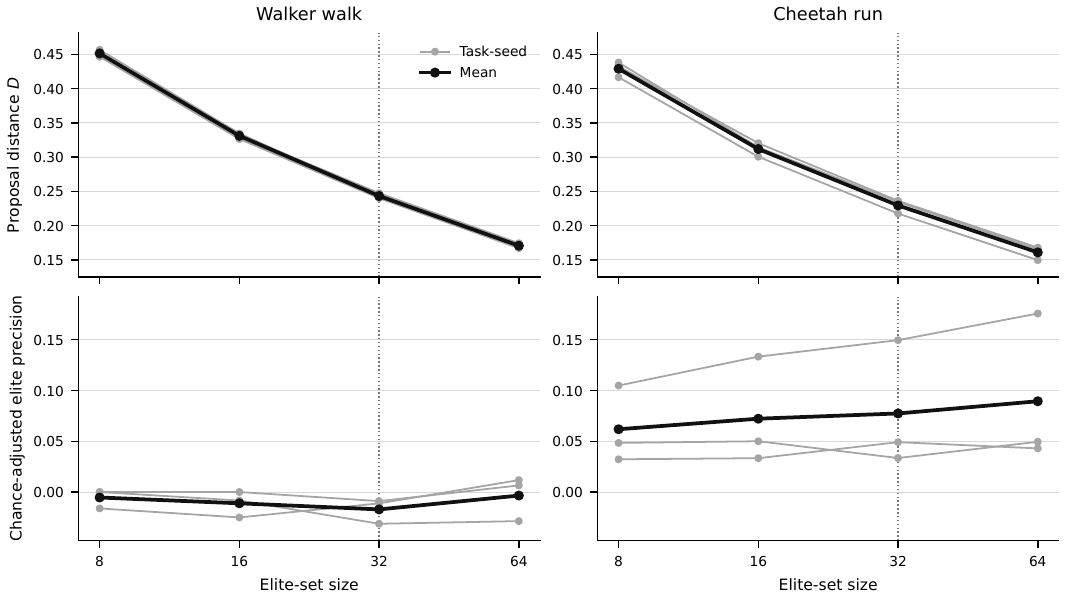}
  \caption{Final-iteration proposal distance and chance-adjusted elite
  precision for 8, 16, 32, and 64 elites on the original seeds 0--2,
  recomputed from the same saved pools.
  Grey lines are task--seed units, black lines are task means, and the dotted
  line marks the fixed cut of 32. Distance decreases as more candidates
  enter the refit; task-mean adjusted precision stays below 0.10 at every cut.}
  \label{fig:app-elite-cuts}
\end{figure}

\section{Clipping diagnostics}
\label{app:clipping}

Candidate actions are clipped to the environment bounds after Gaussian
sampling. Table~\ref{tab:app-bound-effects} records the raw displacement
of the pool mean caused by clipping and the final proportion of proposal
coordinates at the width floor. Figure~\ref{fig:app-clipping} records the
fraction of candidates with at least one clipped coordinate along each
generator trace for the original seeds 0--2. The displacement and clipping rate measure different
effects: many clipped coordinates can produce little mean displacement
when their shifts cancel.

\begin{table}[H]
  \centering
  \begin{tabular}{lcc}
    \toprule
    Measurement & Walker & Cheetah \\
    \midrule
    Initial mean displacement (action units) & \PTWalkerClippingDisplacementRawInitial & \PTCheetahClippingDisplacementRawInitial \\
    Final mean displacement (action units) & \PTWalkerClippingDisplacementRawFinal & \PTCheetahClippingDisplacementRawFinal \\
    Final width-floor occupancy (\%) & \PTWalkerFloorOccupancyFinalPercent & \PTCheetahFloorOccupancyFinalPercent \\
    \bottomrule
  \end{tabular}
  \caption{Bound-related diagnostics for the combined six-seed audit.
  Clipping displacement compares each pool's means before and after clipping,
  in root-mean-square action units. Measurements average model families and
  states within task-seed units, then seeds within tasks. Mean displacement
  decreases, and most final proposal coordinates remain above the width
  floor. The action bounds and floor are fixed across the traces.}
  \label{tab:app-bound-effects}
\end{table}

\begin{figure}[H]
  \centering
  \includegraphics[width=0.98\textwidth]{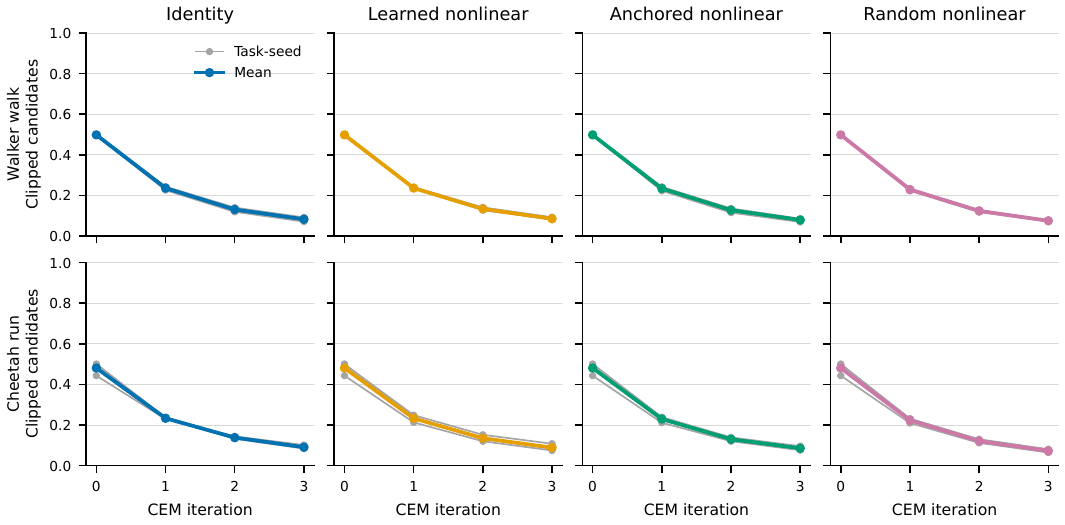}
  \caption{Fraction of candidates with at least one clipped action coordinate,
  separated by task and pool generator for the original seeds 0--2.
  CEM settings and base samples are fixed
  across generators. Every task--generator mean decreases at each update, from
  0.48--0.50 at iteration zero to below 0.10 at iteration three.}
  \label{fig:app-clipping}
\end{figure}

\section{Random-elite reference}
\label{app:random-elites}

The magnitude of $D$ depends on the proposal and the finite candidate pool. As
a reference, we draw 4,096 random subsets of 32 candidates from each saved
pool in the original seeds 0--2, sampling without replacement within each
subset. Fitting each subset produces a pool-specific
distribution of proposal distances from the environment-elite update.

\begin{figure}[H]
  \centering
  \includegraphics[width=0.98\textwidth]{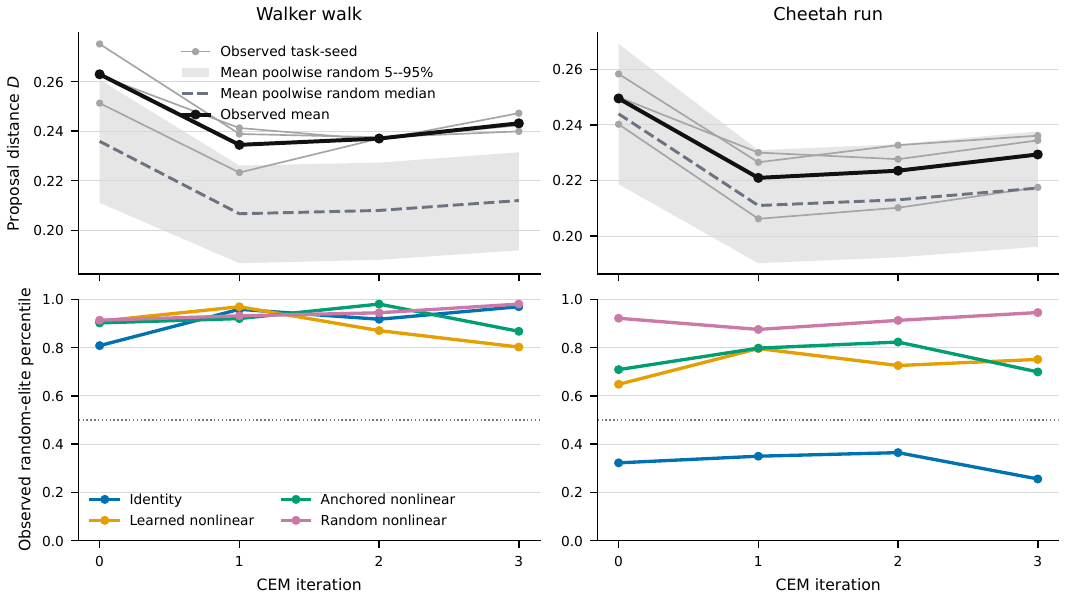}
  \caption{Observed diagonal proposal distances against 32-candidate random
  refits from the same saved pools for the original seeds 0--2.
  The upper row compares task--seed values
  with the mean random median and poolwise 5--95\% interval; the lower row gives
  each surrogate's random-reference percentile. The observed refit lies above
  the random median in 28 of 32 task--surrogate--iteration summaries; Cheetah
  Identity supplies the four exceptions. The reference scales $D$ within a
  finite pool and does not measure realised cost.}
  \label{fig:app-random-elites}
\end{figure}

\section{Reproducibility settings}
\label{app:reproducibility}

\begin{table}[H]
  \centering
  \small
  \setlength{\tabcolsep}{4pt}
  \renewcommand{\arraystretch}{1.06}
  \begin{tabularx}{\textwidth}{@{}>{\raggedright\arraybackslash}p{0.20\textwidth}X@{}}
    \toprule
    Item & Setting \\
    \midrule
    Runtime
      & Python 3.11.3; PyTorch 2.10.0; \texttt{dm\_control} 1.0.37;
        MuJoCo 3.5.0; CUDA 12.8.0. \\
    Dreamer architecture
      & Encoder channels 32--64--128--256; categorical RSSM with deterministic
        width 256 and \(32\times32\) stochastic categories; context width 1280;
        actor and critic hidden width 256. \\
    Dreamer training
      & 500k environment steps per task--seed; five prefill episodes; replay
        capacity 500k; Adam with learning rates \(3\times10^{-4}\) (world model)
        and \(10^{-4}\) (actor, critic), no weight decay; batches of eight
        length-16 sequences; one update per environment step; gradient clip
        100. KL, reconstruction, reward, and continuation weights are 1;
        free nats is 1. The imagination horizon is 15, with
        \(\gamma=0.99\), \(\lambda=0.95\), and actor entropy weight
        \(3\times10^{-4}\). The final \texttt{latest.pt} checkpoint is frozen. \\
    Surrogate data
      & 32 deterministic frozen-actor episodes per task--seed, shuffled with
        NumPy seed 0 and split 26/3/3 for training/validation/holdout; length-32
        sequences; mean latent inference; rollout targets through horizon 15. \\
    Lift networks
      & Identity: width 1280. Learned nonlinear:
        \(1280\!:\!512\!:\!512\!:\!256\). Anchored nonlinear: affine
        \(1280\!:\!96\) concatenated with
        \(1280\!:\!480\!:\!512\!:\!160\). Random nonlinear: fixed
        \(1280\!:\!512\!:\!512\!:\!256\). MLP hidden layers use SiLU; outputs
        are linear. \\
    Shared surrogate layers
      & Transition \(z_{t+1}=Az_t+Ba_t+b\). Compressed-family decoder
        \(256\!:\!512\!:\!512\!:\!1280\); Identity uses the identity decoder.
        Reward and continuation heads are \(d_z\!:\!256\!:\!256\!:\!1\). \\
    Surrogate optimisation
      & AdamW; learning rate \(3\times10^{-4}\); weight decay \(10^{-5}\);
        batch size 128; gradient clip 100; 100 epochs without early stopping.
        Validation runs each epoch; the retained checkpoint minimises validation
        decoded-context prior-rollout loss. \\
    Surrogate loss
      & Horizons \(\{1,3,5,8,15\}\), each weighted 1. Latent-state rollout has
        weight 1; the summed posterior and prior decoded-context terms have
        weight 1. Reward MSE and continuation binary cross-entropy have weight
        0.25 each. Direct context reconstruction, variance, and covariance terms
        have weight 0. \\
    Task readout
      & Linear ridge regression with intercept from decoded context to task
        variables; coefficient \(10^{-3}\), intercept unregularised. Four
        deterministic actor episodes use seeds 120000--120003. \\
    Planning cost
      & Discount \(\gamma=1\), terminal scale \(\lambda_T=2.5\), action weight
        \(R=0.01I_6\), and actor-deviation weight 0. Predicted and realised costs
        use the same weighted squared tracking error. \\
    Elite and pairwise ties
      & Elite ties use candidate index order. Pairwise agreement excludes
        realised-cost differences of at most \(10^{-12}\); a model tie
        against a strict realised ordering receives no agreement credit. \\
    Walker weights
      & \(Q=\operatorname{diag}([8,1^{\times14},2^{\times9}])\) on height,
        orientations, and velocity, against the fixed actor-visited target. \\
    Cheetah weights
      & \(Q=\operatorname{diag}([1^{\times8},3^{\times9}])\) on position and
        velocity, against the matched frozen-actor reference trajectory. \\
    \bottomrule
  \end{tabularx}
  \caption{Settings needed to reproduce the frozen models, surrogate fits,
  readouts, and costs used by the audit. Architecture strings list layer
  widths from input to output.}
  \label{tab:reproducibility-settings}
\end{table}

\paragraph{Compute required for reproduction.}
Training the twelve backbones, fitting the 48 surrogates and running the
reported audits and splices requires approximately 179 GPU-hours in total.
The breakdown is 133 hours for the twelve 500k-step Dreamer backbones,
4.6 hours for data exports and surrogate fits, and 41 hours for the crossed
audit and twelve splice units. The first two estimates allocate $12/48$ of
the corresponding 48-unit training budgets to the units used here.
The audit and splice use four CPU cores, 32 GB RAM and one NVIDIA L40S
per unit. Walker and Cheetah crossed units take approximately 1.5 and
3.1 hours, respectively. The six held-out splice units account for
18,432 candidate executions and 2.6 of the 41 GPU-hours.

These figures sum wall-clock device time on H200, H100, A100, L40S and
V100 GPUs without adjusting for their different speeds. They estimate the
cost of reproducing the reported experiment and exclude unused tasks,
other evaluations, failed jobs and exploratory runs.

\paragraph{Code and data.}
The code, trained checkpoints and saved simulator states are not public
with this preprint. We plan to release the code and reproduction
instructions with an archival publication.

\fi

\end{document}